\documentclass[conference]{IEEEtran}
\IEEEoverridecommandlockouts
\usepackage{cite}
\usepackage{amsmath,amssymb,amsfonts}
\usepackage{algorithm,algpseudocode}
\usepackage{graphicx}
\usepackage{textcomp}
\usepackage{xcolor}
\usepackage{multirow}
\def\BibTeX{{\rm B\kern-.05em{\sc i\kern-.025em b}\kern-.08em
    T\kern-.1667em\lower.7ex\hbox{E}\kern-.125emX}}
\begin{document}

\title{Differential Evolution-based Neural Network Training Incorporating a Centroid-based Strategy and Dynamic Opposition-based Learning\\
}

\author{\IEEEauthorblockN{Seyed Jalaleddin Mousavirad}
	\IEEEauthorblockA{\textit{Department of Computer Engineering} \\
		\textit{Hakim Sabzevari University}\\
		Sabzevar, Iran} %\\
	%Jalalmoosavirad@gmail.com}
	\and
	\IEEEauthorblockN{Diego Oliva, Salvador Hinojosa}
	\IEEEauthorblockA{\textit{Depto. de Ciencias Computacionales} \\
		\textit{Universidad de Guadalajara}\\
		Guadalajara, Mexico}% \\
	\and
	\IEEEauthorblockN{Gerald Schaefer}
	\IEEEauthorblockA{\textit{Department of Computer Science}\\
		\textit{Loughborough University}\\
		Loughborough, U.K.}\\
}

\IEEEoverridecommandlockouts

\IEEEpubid{\makebox[\columnwidth]{978-1-7281-8393-0/21/\$31.00~\copyright2021 IEEE \hfill} %--> insert the copyright option applicable from above.
	\hspace{\columnsep}\makebox[\columnwidth]{ }}
\maketitle
\IEEEpubidadjcol

\begin{abstract}
Training multi-layer neural networks (MLNNs), a challenging task, involves finding appropriate weights and biases. MLNN training is important since the performance of MLNNs is mainly dependent on these network parameters. However, conventional algorithms such as gradient-based methods, while extensively used for MLNN training, suffer from drawbacks such as a tendency to getting stuck in local optima. Population-based metaheuristic algorithms can be used to overcome these problems. In this paper, we propose a novel MLNN training algorithm, CenDE-DOBL, that is based on differential evolution (DE), a centroid-based strategy (Cen-S), and dynamic opposition-based learning (DOBL). The Cen-S approach employs the centroid of the best individuals as a member of population, while other members are updated using standard crossover and mutation operators. This improves exploitation since the new member is obtained based on the best individuals, while the employed DOBL strategy, which uses the opposite of an individual, leads to enhanced exploration. Our extensive experiments compare CenDE-DOBL to 26 conventional and population-based algorithms and confirm it to provide excellent MLNN training performance.  
\end{abstract}


\section{Introduction}
\label{sec:Intro}
Artificial neural networks (ANNs) are popular pattern recognition techniques to deal with complicated classification and regression problems in various domains, including food quality~\cite{Rice_Jalal01,Rice_Jalal02}, medicine~\cite{ANN_Hasan,ANN_Medical02}, and business~\cite{ANN_Financial01}.

Multi-layer neural networks (MLNNs), which are extensively employed, generally have three types of layers, input, hidden, and output layers, while each connection has a specific weight that ascertains its strength. Training an MLNN means finding appropriate weights for the connections, and is a challenging and important task since the performance of an MLNN is directly related to these parameters~\cite{ANN_Benchmark_mine}.

Gradient-based approaches such as back-propagation are widely used for MLNN training, but have drawbacks such as being sensitive to the initial weights and a tendency to get stuck in local optima. Population-based metaheuristic (PBMH) algorithms such as particle swarm optimisation (PSO)~\cite{PSO_Main_Paper}, differential evolution~\cite{DE_Original}, and human mental search (HMS)~\cite{HMS_Main_Paper,HMS_Main_Paper2} are capable of overcoming these problems. PBMHs are problem-independent optimisation algorithms that find an optimal solution using a population of candidate solutions and some specific operators. Nowadays, these algorithms are extensively employed for MLNN training due to their simplicity, flexibility, and ability to escape local optima. PBMH-based training algorithms have been introduced using particle swarm optimisation (PSO)~\cite{ANN_PSO01,ANN_BLS,ANN_CLPSO}, artificial bee colony (ABC)~\cite{ANN_ABC01}, imperialist competitive algorithm (ICA)~\cite{ANN_ICA01,ANN_ICA_Memetic,ANN_Training_ISNN}, firefly algorithm (FA)~\cite{ANN_FA01}, grey wolf optimiser (GWO)~\cite{ANN_GWO01,ANN_GWO_Mine}, ant lion optimiser~\cite{ANN_ALO01}, dragonfly algorithm (DA)~\cite{ANN_DA01}, sine cosine algorithm~\cite{ANN_SCA01}, whale optimisation algorithm (WOA)~\cite{ANN_WOA01}, grasshopper optimisation algorithm~\cite{ANN_GOA01}, and salp swarm algorithm (SSA)~\cite{ANN_SSA01}, among others.

Differential evolution (DE)~\cite{DE_Original} is an effective PBMH algorithm with demonstrated excellent performance in solving complex optimisation problems~\cite{DE_Deceptive_ICCSE2019,DE_Clustering01,DE_Competition_ICCSE2019}. DE is based on mutation, crossover, and selection operators, where mutation is responsible for generating a mutant individual based on scaled differences among individuals, crossover combines the mutant individual with the parent one, and selection carries over the better individuals to the next population.

DE has shown satisfactory performance in finding optimal weights in MLNNs. \cite{ANN_DE01} proposes a DE-based algorithm and compares it with gradient-based algorithms, indicating DE to achieve higher accuracy, while~\cite{ANN_DE03} employs a DE algorithm with multiple trial vectors for MLNN training. \cite{ANN_DE_opposition} proposes a novel training algorithm based on quasi-opposition-based learning, showing the improved DE to obtain better performance in classification problems. Recently, \cite{ANN_RDE-OP} introduces a region-based DE algorithm combined with quasi-opposition-based learning, RDE-OP, for MLNN training. RDE-OP benefits from a clustering algorithm to partition the current population so that each cluster centre acts as a crossover operator. 

In this paper, we propose a novel DE-based training algorithm, CenDE-DOBL, that is based on a centroid-based strategy incorporating opposition-based learning. In our proposed algorithm, a centroid individual is injected into the current population. Since the centroid individual is based on the best individuals, this improves the exploitation ability of the algorithm. On the other hand, dynamic opposition-based learning (DOBL) is incorporated to further enhance the exploration ability of the algorithm. We perform an extensive set of experiments, comparing CenDE-DOBL with 26 conventional and population-based algorithms and demonstrate it to give excellent MLNN training performance and to outperform the other methods.

The remainder of the paper is organised as follows. Section~\ref{sec:back} describes some background on differential evolution and opposition-based learning. Section~\ref{sec:proposed} then details our proposed CenDE-DOBL algorithm, while experimental results are presented in Section~\ref{sec:exp}. Finally, Section~\ref{sec:concl} concludes the paper.

\section{Background}
\label{sec:back}
\subsection{Differential Evolution}
Differential evolution (DE)~\cite{DE_Original} is an effective PBMH algorithm which has shown remarkable performance in solving different complex optimisation problems~\cite{Image_Segmentation_auto_DE,Image_Thresholding_CEC2019,DE_Application01}. DE commences with $N_{P}$ individuals generated by a uniform distribution, and has three main operators: mutation, crossover, and selection. Mutation creates a vector, called mutant vector, as  
\begin{equation}
v_{i}=x_{r1}+F*(x_{r2}-x_{r3}),
\end{equation} 
where $x_{r1}$, $x_{r2}$, and $x_{r3}$ are three distinct randomly selected individuals, and $F$ is a scaling factor. 

Crossover combines the mutant and target vectors. A popular crossover operator is binomial crossover, defined as
\begin{equation}
u_{i,j}=\begin{cases}
v_{i,j} & rand(0,1)\leq CR \text{ or } j==j_{rand} \\
x_{i,j} & \text{otherwise}
\end{cases} ,
\end{equation}
where $i=1,..., NP$, $j=1,..., D$, $CR$ is the crossover rate, and $j_{rand}$ is a random number in $[1;NP]$. 

Finally, the selection operator is responsible for choosing the better individual from the trial and target vectors.

\subsection{Opposition-based Learning}
Opposition-based learning (OBL)~\cite{OBL_Tizhoosh} aims to yield improved performance by employing opposition individuals. Assume that $x = [x_{1},x_{2},...,x_{N}]$ is a number in an $N$-dimensional space. The corresponding opposite number of $x$ is then defined as  
\begin{equation}
\label{OBL_DF}
\check{x_{i}}=a_{i}+b_{i}-x_{i} ,
\end{equation}
where $a_{i}$ and $b_{i}$ are the lower and upper bounds of the search space.

Dynamic quasi-opposition-based learning (DOBL)~\cite{OBL_Quasi_ODE} is a variant of OBL employing quasi-opposition numbers, and is dynamic since the maximum and minimum values of the individuals are employed to create an opposite individual. \cite{OBL_Quasi_ODE} shows that in a black-box optimisation quasi-opposition numbers have a higher chance of getting closer to the optimum in comparison to opposite numbers. The quasi-opposite number of $x$ is obtained as   
\begin{equation}
\label{DOBL_DF}
\check{x_{i}}=rand[\frac{( a_{i}+b_{i})}{2},(a_{i}+b_{i}-x_{i}) ] ,
\end{equation}
where $rand[m,n]$ is a uniform random number between $m$ and $n$.

\section{CenDE-DOBL Algorithm}
\label{sec:proposed}
Neurons are the main components of MLNNs. While every neuron performs a minor task, their co-operation enables a neural network to handle complex pattern recognition tasks. In general, an MLNN has three types of layers, an input layer, hidden layers, and an output layer. Each connection between two neurons has a specific weight that determines the strength of each connection, while neurons also have bias terms. MLNN training means finding appropriate weights and biases and is a challenging task.

Our proposed CenDE-DOBL algorithm for MLNN training is based on a combination of a centroid-based strategy and a dynamic opposition-based learning strategy with the former improving exploitation and the latter exploration. In the following, we first describe the main components of CenDE-DOBL and then explain how these fit together to form the algorithm.

\subsection{Centroid-based Strategy}
Centre-based sampling is a concept based on the centroid of individuals to improve a metaheuristic algorithm~\cite{center_sampling_PSO_SMC2020,DE_Deceptive_ICCSE2019}. \cite{center_sampling_01} shows that, based on Monte-Carlo simulation, the probability of individuals being closer to an unknown solution is higher towards the centre of the search space compared to randomly-located individuals.

% no point of including this since it's not being used (only the centroid of individuals is being used)
%For a search space in the range of $[a_i,b_i]$ for the $i$-th dimension, the centroid is calculated as \begin{equation}
%c_{i}=\frac{a_{i}+b_{i}}{2},
%\end{equation} 
%where $i=1,...,D$ and $D$ is dimensionality of the search space.

Inspired by~\cite{center_DE_Mine01}, CenDE-DOBL benefits from a centroid-based individual that is created based on the $N$ best individuals. In our proposed algorithm, all individuals except one are updated based on standard operators, while the last individual is the centroid of the $N$ best individuals, defined as
\begin{equation}
\label{Eq:cen}
\overrightarrow{x_{center}}=\frac{\overrightarrow{x_{b1}}+...\overrightarrow{x_{bi}}+...+\overrightarrow{x_{bN}}}{N} ,
\end{equation}
where $\overrightarrow{x_{bi}}$ is the $i$-th best individual.

% not useful as trivial
%\begin{figure}
%	\centering
%	\begin{subfigure}[b]{0.45\textwidth}
%		\centering
%		\includegraphics[width=\textwidth]{Figures/Centroid_Pop.png}		
%	\end{subfigure}	
%	\caption{$N_{P}-1$ individuals update their position based on the standard mutation and crossover, while the last individual is responsible for keeping the center of the $N$ best individuals.} 
%	\label{Centroid_Pop}
%\end{figure}

\begin{figure}[b!]
	\centering
	\includegraphics[width=.9\columnwidth]{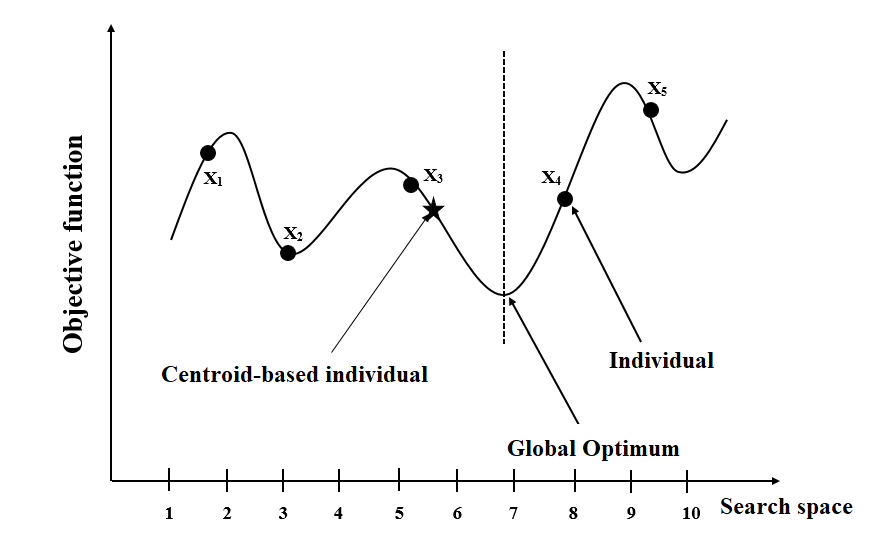}
	\caption{Visualisation of centroid-based candidate solution for a one-dimensional problem.} 
	\label{fig:center}
\end{figure}

Fig.~\ref{fig:center} visualises the concept for a 1-D problem with $N_{P}=6$ where 5 individuals are created using the standard operators, and the three best individuals, $\overrightarrow{x_{2}}$, $\overrightarrow{x_{3}}$, and $\overrightarrow{x_{4}}$ with positions at $\{3,5,8\}$ are used to create the centroid-based individual $\overrightarrow{x_{center}}$ at 5.33.

It is worthwhile to mention that (1) the centroid-based individual does not require any additional function evaluations, and (2) the centroid-based individual is based on the best individuals and, therefore, enhances exploitation.

%\begin{algorithm}[t!]
%	\caption{Pseudo-code of centroid-based strategy for generating an individual.}
%	\label{Alg1:centroid}
%	\small
%	\begin{algorithmic}[1]
%		\Procedure{Cen-S Algorithm}{}
%		\State // Variables: $L$: lower bound; $U$: upper bound,  $N_{P}$: population size
%		\State \
%		\State $Pop(N_{P})=\frac{\overrightarrow{x_{b1}}+\overrightarrow{x_{b2}}+...+\overrightarrow{x_{bN}}}{N}$

%		\EndProcedure
%	\end{algorithmic}
%\end{algorithm}

\subsection{Opposition-based Strategy}
CenDE-DOBL uses an OBL scheme in two ways, for initialisation and for generating jumps. Algorithm~\ref{Alg1:OBL} shows the employed OBL strategy in form of pseudo-code.
%Quasi-opposition-based initialisation and dynamic quasi-opposition-based generating jumping algorithms are called \textit{QO-Ini} and \textit{DOQ-GR} algorithms, respectively.

\begin{algorithm}[h!]
	\caption{Pseudo-code of OBL($D$, $N_{P}$, $Pop$ $L$, $U$) algorithm}
	\label{Alg1:OBL}
	\small
	\begin{algorithmic}[1]
		\Procedure{\textit{OBL}}{}
		\State // Variables: $D$: dimensionality, $N_{P}$: population size, $Pop$: initial population, $L$: lower bound, $U$: upper bound
		\State \
		\For {$i$ from 0 to $N_{P}$} 
		\For {$j$ from 0 to $D$}
		\State $\check{OPop(i,j)}=
		rand[\frac{(L(i,j)+U(i,j)}{2},(L_{i,j}+U_{i,j}-Pop(i,j))]$
		\EndFor
		
		\EndFor
		
		\State Evaluate objective function value for each individual based on Eq.~(\ref{eq:obj})
		\State $Pop\leftarrow$ Select $N_{P}$ best individuals from set $\{Pop,OPop\}$ as initial population
		\EndProcedure
	\end{algorithmic}
\end{algorithm} 

After the initial population is generated, a quasi-opposition-based population ($OPop$) is generated using Eq.~(\ref{DOBL_DF}). We then select the $N_{P}$ best individuals from the union of the initial population and the opposition-based population.
%QO-Ini algorithm in the form of pseudo-code can be seen in Algorithm~\ref{alg2}.

After generating new individuals using mutation and crossover operators, the proposed algorithm generates a quasi-opposition-based population based on a jumping rate $J_{r}$ between 0 and 0.4~\cite{OBL_Quasi_ODE}. Then, a new population is generated based on the best individuals from the current population and the quasi-opposition-based population. The OBL strategy in this step is dynamic since the maximum and minimum values of the individuals are employed to create an opposite individual as 
\begin{equation}
\textstyle \check{x}_{i,j}=\min_{j}^{p}+\max_{j}^{p}-x_{i,j}  \thickspace
i=1,2,...,N_{p} \thickspace
j=1,2,...,D ,
\end{equation}
where $\min_{j}^{p}$ and $\max_{j}^{p}$ indicate the minimum and maximum of the population in the $j$-th dimension.

%\begin{algorithm}[t!]
%	\caption{Pseudo-code of \textit{DQO-GR} algorithm}
%	\label{Alg1:centroid}
%	\small
%	\begin{algorithmic}[1]
%		\Procedure{\textit{DQO-GR} algorithm}{}
%		\State // Variables: $D$: Number of dimensions, $N_{P}$: Population size, $J_{r}$: Jumping rate, $Pop$: Current population, $L$: Lower bound, $U$: Upper bound
%		\State \
%		\If {$rand(0,1)<J_{r}$}
%		\For {$i$ from 0 to $N_{P}$} 
%		\For {$j$ from 0 to $D$}
%		\State $\check{OPop(i,j)}=
%		rand[\frac{(L(i,j)+U(i,j)}{2},(L_{i,j}+U_{i,j}-Pop(i,j))]$
%		\EndFor

%		\EndFor
%		\EndIf

%		\State 	Evaluate the objective function for each individual in based on Eq.~\ref{eq:obj}
%		\State $Pop\leftarrow$ Select $N_{P}$ best individuals from set of $(Pop,OPop)$ as the initial population.

%		\EndProcedure
%	\end{algorithmic}
%\end{algorithm} 

\subsection{DE/local-to-best/1 Strategy}
Instead of the standard mutation operator, we employ a \textit{DE/local-to-best/1} strategy in which the base vector is a combination of one randomly-selected individual and the best individual of the previous population as
\begin{equation}
v_{i}=x_{i}+F*(x_{best}-x_{i})+F*(x_{r2}-x_{r3}),
\end{equation} 
where $x_{i}$ and $x_{best}$ are the $i$-th and the best member of the old population,  $x_{r1}$ and $x_{r2}$ are two different randomly-selected individuals, and $F$ is a scaling factor. This approach tries to strike a balance between robustness and fast convergence.

\subsection{Encoding Strategy}
\label{sec:encode}
\begin{figure}[b!]
	\centering
	\includegraphics[width=.9\columnwidth]{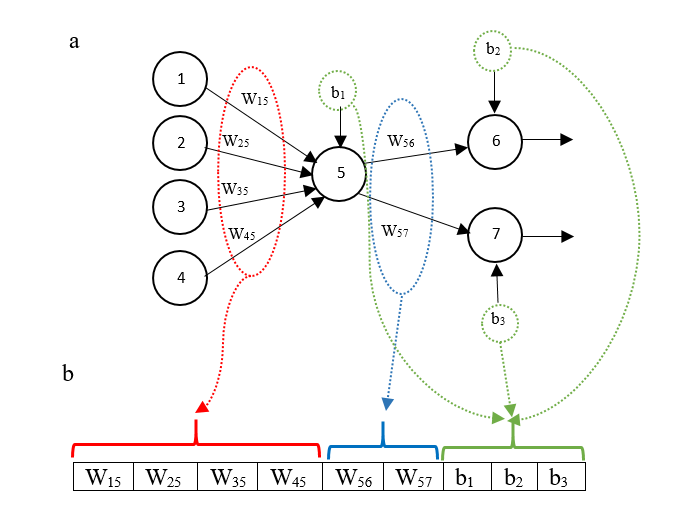}
	\caption{Illustration of encoding  strategy. Top: network, bottom: resulting structure of individual.} 
	\label{fig:encoding}
\end{figure}

CenDE-DOBL employs a real-valued encoding strategy to encode the connection weights and biases.  Consequently, each individual's length is equal to the total number of weights and bias terms. Fig.~\ref{fig:encoding} illustrates the encoding strategy for a sample MLNN with one neuron in the single hidden layer.

\subsection{Objective Function}
\label{sec:obj}
We use an objective function based on classification error defined as
\begin{equation}
E = \frac{100}{P} \sum_{p=1}^{P} \xi(x_p),
\label{eq:obj}
\end{equation}
with
\begin{equation}
\xi(x_p) =
\begin{cases}
1 & \text{if } o_{p} \neq d_{p} \\
0 & \text{otherwise}
\end{cases} ,
\label{eq7}
\end{equation}
where $d_{p}$ and $o_{p}$ are the desired and predicted output, respectively, of input $x_p$, the $p$-th of $P$ test samples. The aim of CenDE-DOBL is to find weights and biases so as to minimise the classification error.

\subsection{Algorithm}
While OBL increases the exploration of the algorithm, the centroid-based strategy enhances its exploitation. Switching between these two schemes is based on a probability, meaning that in each iteration, only one of these is performed. The whole CenDE-DOBL algorithm is given in Algorithm~\ref{Alg:proposed} in the form of pseudo-code.

\begin{algorithm}[h!]
	\caption{Pseudo-code of CenDE-DOBL($D$, $Max_{NFC}$, $N_{P}$, $J_{r}$, $N$, $L$, $U$) algorithm for MLNN training.}
	\label{Alg:proposed}
	\small
	\begin{algorithmic}[1]
		\Procedure{\textit{CenDE-DOBL}}{}
		\State // Variables:$D$: dimensionality, $Max_{NFC}$: maximum number of function evaluations, $N_{P}$: population size,  $J_{r}$: jumping rate, $N$: number of best solutions, $L$: lower bound, $U$: upper bound 
		\State \
		\State Generate initial population $Pop$ randomly based on the encoding strategy introduced in Section~\ref{sec:encode}\;
		\State Call OBL($D$, $N_{P}$, $Pop$ $L$, $U$) algorithm for generating a new population based on DOBL strategy
		\State Calculate objective function value for each individual based on Eq.~(\ref{eq:obj})
		\While {$NFE<=NFE_{\max}$}
		\State Select three parents, $x_{r1}$ and $x_{r2}$, randomly from the current population, with $x_{r1} \neq x_{r2} $
		\State $v_{i}=x_{i}+F*(x_{best}-x_{i})+F*(x_{r1}-x_{r2})$
		\For {$j$ from 0 to $D$}
		\If {$rand_{j}[0,1]<C_{R}\thickspace or \thickspace j==j_{rand}$}
		\State $u_{i,j}={v}_{i,j}$
		\Else
		\State $u_{i,j}=x_{i,j}  $
		\EndIf
		\EndFor
		\State 	Calculate objective function value of $u_{i}$ based on Eq.~(\ref{eq:obj})
		\If {$f(u_{i})<f(x_{i})$}
		\State $ \bar{x} \leftarrow u_{i} $
		\Else
		\State $\bar{x} \leftarrow x_{i}$
		\EndIf
		\State
		\If {$rand(0,1)<J_{r}$}
		\State $L_{New}=$ minimum of all individuals
		\State $U_{New}=$ maximum of all individuals
		\State Call OBL($D$ , $N_{P}$, $Pop$, $L_{New}$, $U_{New}$) algorithm
		\Else
		\State Select $N$ best individuals
		\State $Pop(N_{P})=\frac{\overrightarrow{x_{b1}}+\overrightarrow{x_{b2}}+...+\overrightarrow{x_{bN}}}{N}$
		\EndIf
		\EndWhile
		\State $x^{*}\leftarrow$ the best individual in the population
		\EndProcedure
	\end{algorithmic}
\end{algorithm}

\section{Experimental Results}
\label{sec:exp}
To assess our proposed CenDE-DOBL algorithm, we conduct a set of experiments with different datasets from different domains with diverse characteristics from the UCI machine learning repository\footnote{https://archive.ics.uci.edu/ml/index.php}, namely
\begin{itemize}
	\item 
	\textit{Iris}: this dataset is one of the most commonly used datasets in the literature. It includes 150 instances, 4 features, and 3 classes. One class is linearly separable from two others, while the latter are not linearly separable. 
	\item
	\textit{Breast Cancer}: this dataset contains 699 instances placed in 2 classes with 9 features such as menopause and tumour size.
	\item
	\textit{Liver}: this clinical dataset from BUPA Medical Research Ltd. has 345 samples, 2 classes and 7 features. %Some features are alkphos alkaline phosphotase and sgot aspartate aminotransferase.
	\item
	\textit{Pima}: this binary classification dataset is a challenging problem with 768 instances and 8 features.
	\item
	\textit{Seed}: this agricultural dataset includes seven geometrical properties of kernels such as compactness, perimeter, and area belonging to three distinct wheat classes with 210 instances. 
	\item
	\textit{Vertebral}: this clinical dataset includes biomechanical features such as pelvic incidence and pelvic tilt employed to classify orthopaedic patients into 3 classes, normal, disk hernia, and spondylolysthesis.
\end{itemize} 

Since our paper does not focus on the best MLNN structure, we follow~\cite{ANN_Training_ISNN,ANN_BLS} and set the number of neurons in the hidden layer to $2 D +1$ where $D$ is the number of input features. Therefore, the number of connection weights for Iris, Cancer, Liver, Pima, Seed, and Vertebral datasets are 43, 210, 105, 171, 136, and 105, respectively. We use $k$-fold cross-validation, with $k=10$, for evaluation, where the dataset is divided into $k$ folds, one fold for testing and the others for training. This process is repeated $k$ times so that each fold is employed once as test data.

We compare CenDE-DOBL with an extensive set of algorithms including both conventional and population-based algorithms. The number of function evaluations for all population-based algorithms is set to 25,000, similar to the number of iterations for all conventional algorithms. The population size for all population-based algorithms is set to 50. For CenDE-DOBL, the crossover probability, scaling factor, number of best individuals, and jumping rate are set to 0.9, 0.5, 3, and 0.3, respectively. For the other algorithms, we employ default parameters values from the cited publications.

In the first experiment, we compare our algorithm with DE, QODE~\cite{ANN_DE_opposition}, and RDE-OP~\cite{ANN_RDE-OP}. We select DE since our proposed algorithm is based on DE, and QODE and RDE-OP because they are among the most recent DE-based training algorithms. Table~\ref{Result_DE} indicates the results in terms of mean and standard deviation as well as their ranking and the resulting average rank.

\begin{table*}[]
	\centering
	\caption{10CV classification accuracy for all datasets for DE, QODE, and RDE-OP in comparison to CenDE-DOBL.}
	\label{Result_DE}
	\begin{tabular}{l|c|cccccc|c}
		&  & Iris & Cancer & Liver & Pima & Seed & Vertebral & avg. rank \\ \hline
		\multirow{3}{*}{DE} & mean & 92.00 & 97.36 & 67.81 & 76.94 & 70.00 & 85.16 &  \\
		& stddev & 5.26 & 2.06 & 8.21 & 4.97 & 11.01 & 5.31 &  \\
		& rank & 4 & 4 & 4 & 4 & 2 & 4 & 3.67 \\ \hline
		\multirow{3}{*}{QODE} & mean & 95.33 & 98.10 & 76.82 & 79.55 & 67.62 & 88.39 &  \\
		& stddev & 6.32 & 0.99 & 9.46 & 4.95 & 3.01 & 8.76 &  \\
		& rank & 3 & 3 & 2 & 3 & 3.5 & 1.5 & 2.58 \\ \hline
		\multirow{3}{*}{RDE-OP} & mean & 96.67 & 98.82 & 75.63 & 80.21 & 67.62 & 86.77 &  \\
		& stddev & 6.48 & 1.67 & 6.45 & 5.73 & 4.92 & 4.42 &  \\
		& rank & 2.00 & 1.00 & 3.00 & 2.00 & 3.50 & 3.00 & 2.42 \\ \hline
		\multirow{3}{*}{CenDE-DOBL} & mean & 98.67 & 98.68 & 78.79 & 81.90 & 90.95 & 88.39 &  \\
		& stddev & 2.81 & 1.08 & 8.64 & 3.17 & 10.15 & 5.09 &  \\
		& rank & 1 & 2 & 1 & 1 & 1 & 1.5 & 1.33 \\ \hline
	\end{tabular}
\end{table*}

As we can see from there, CenDE-DOBL gives the best results for 5 of the 6 cases and is ranked second for the remaining one. On the Iris dataset, our algorithm obtains the first rank with a classification accuracy improvement of 2\% or more compared to the other algorithms. On the Cancer dataset, RDE-OP gives slightly better results, by 0.14\%, than CenDE-DOBL. CenDE-DOBL outperforms the other algorithms by over 1.9\% on the Liver dataset, while for the Pima dataset, DE, QODE, and RDE-OP achieve accuracies of 76.94\% , 67.62\%, and 67.62\%, respectively in comparison to 81.90\% for CenDE-DOBL. An even greater improvement can be seen for the Seed dataset, where CenDE-DOBL obtains a mean accuracy of 90.95\%, while the next-best algorithm (DE) yields only 70.00\%. Finally, on the Vertebral dataset, CenDE-DOBL and  QODE are tied to give the best results, however the standard deviation for CenDE-DOBL is smaller, indicating more robust performance.

In the next experiment, we compare CenDE-DOBL with 12 conventional algorithms, namely gradient descent with momentum backpropagation (GDM)~\cite{ANN_GDM}, gradient descent with adaptive learning rate backpropagation (GDA)~\cite{ANN_Hagan}, gradient descent with momentum and adaptive learning rate backpropagation (GDMA)~\cite{ANN_Book_conv}, conjugate gradient backpropagation with Fletcher-Reeves updates (CG-FR)~\cite{ANN_CG_FR}, conjugate gradient backpropagation with Polak-Ribiere updates (CG-PR)~\cite{ANN_CG_FR_01,ANN_CG_FR_02}, conjugate gradient backpropagation with Powell-Beale restarts (CG-PBR)~\cite{ANN-CG-PBR}, BFGS quasi-Newton backpropagation (BFGS)~\cite{ANN_BFGS}, Levenberg-Marquardt backpropagation (LM)~\cite{Levenberg01,Levenberg02}, one-step secant backpropagation (OSS)~\cite{ANN_OSS}, resilient backpropagation (RP)~\cite{ANN_RP}, scaled conjugate gradient backpropagation (SCG)~\cite{ANN_SCG}, and Bayesian regularisation backpropagation(BR)~\cite{ANN_BR}.

The results are given in Table~\ref{Result_Conv}. In all cases, our proposed algorithm gives the highest classification accuracy (once tied with BR), thus outperforming all other methods by a wide margin. 

\begin{table*}[]
	\centering
	\caption{10CV classification accuracy for all datasets for conventional ANN training algorithms in comparison to CenDE-DOBL.}
	\label{Result_Conv}
	\begin{tabular}{l|c|cccccc|c}
		&  & Iris & Cancer & Liver & Pima & Seed & Vertebral & avg. rank \\ \hline
		\multirow{3}{*}{GDM} & mean & 92.00 & 92.99 & 59.47 & 67.98 & 47.62 & 76.13 &  \\
		& stddev & 8.20 & 7.60 & 15.05 & 13.03 & 29.95 & 9.15 &  \\
		& rank & 12 & 11 & 12 & 13 & 13 & 12 & 12.17 \\ \hline
		\multirow{3}{*}{GDA} & mean & 94.67 & 95.90 & 58.24 & 75.91 & 82.38 & 80.65 &  \\
		& stddev & 5.26 & 2.04 & 6.53 & 4.52 & 6.75 & 5.27 &  \\
		& rank & 10 & 10 & 13 & 9 & 9.5 & 10 & 10.25 \\ \hline
		\multirow{3}{*}{GDMA} & mean & 82.67 & 90.47 & 60.66 & 73.20 & 80.00 & 72.90 &  \\
		& stddev & 16.98 & 5.94 & 13.99 & 6.74 & 16.47 & 17.62 &  \\
		& rank & 13 & 13 & 11 & 12 & 12 & 13 & 12.33 \\ \hline
		\multirow{3}{*}{CG-FR} & mean & 95.33 & 96.19 & 60.86 & 76.69 & 86.67 & 83.87 &  \\
		& stddev & 4.50 & 1.72 & 8.29 & 6.20 & 9.73 & 7.60 &  \\
		& rank & 7.5 & 7 & 10 & 5 & 5.5 & 4 & 6.50 \\ \hline
		\multirow{3}{*}{CG-PR} & mean & 96.00 & 97.07 & 65.18 & 75.12 & 85.24 & 80.97 &  \\
		& stddev & 6.44 & 1.39 & 8.47 & 3.57 & 9.90 & 4.92 &  \\
		& rank & 4 & 2 & 8 & 11 & 8 & 8.5 & 6.92 \\ \hline
		\multirow{3}{*}{CG-PBR} & mean & 94.00 & 96.49 & 67.18 & 76.04 & 88.10 & 82.90 &  \\
		& stddev & 5.84 & 2.95 & 9.09 & 5.06 & 7.86 & 6.09 &  \\
		& rank & 11 & 5 & 3 & 7 & 3.5 & 6 & 5.92 \\ \hline
		\multirow{3}{*}{BFGS} & mean & 95.33 & 96.92 & 65.22 & 76.70 & 86.67 & 84.19 &  \\
		& stddev & 4.50 & 1.28 & 7.06 & 3.11 & 9.73 & 5.98 &  \\
		& rank & 7.5 & 3 & 7 & 4 & 5.5 & 3 & 5.00 \\ \hline
		\multirow{3}{*}{LM} & mean & 96.67 & 96.04 & 65.55 & 76.04 & 88.10 & 83.55 &  \\
		& stddev & 4.71 & 2.51 & 10.44 & 4.66 & 7.53 & 7.04 &  \\
		& rank & 2 & 9 & 6 & 8 & 3.5 & 5 & 5.58 \\ \hline
		\multirow{3}{*}{OSS} & mean & 95.33 & 96.34 & 64.89 & 76.58 & 86.67 & 80.97 &  \\
		& stddev & 4.50 & 2.21 & 8.01 & 4.30 & 5.85 & 8.93 &  \\
		& rank & 7.5 & 6 & 9 & 6 & 7 & 8.5 & 7.33 \\ \hline
		\multirow{3}{*}{RP} & mean & 95.33 & 96.05286 & 65.82 & 76.83 & 80.48 & 78.39 &  \\
		& stddev & 5.49 & 2.47 & 5.21 & 4.50 & 9.38 & 5.05 &  \\
		& rank & 7.5 & 8 & 5 & 3 & 11 & 11 & 7.58 \\ \hline
		\multirow{3}{*}{SCG} & mean & 96.00 & 96.63 & 66.97 & 78.52 & 82.38 & 82.26 &  \\
		& stddev & 8.43 & 2.09 & 9.60 & 3.05 & 9.54 & 8.50 &  \\
		& rank & 4 & 4 & 4 & 2 & 9.5 & 7 & 5.08 \\ \hline
		\multirow{3}{*}{BR} & mean & 96.00 & 91.81 & 70.71 & 75.89 & 90.95 & 84.52 &  \\
		& stddev & 7.17 & 3.82 & 6.93 & 5.37 & 7.60 & 6.94 &  \\
		& rank & 4 & 12 & 2 & 10 & 1.5 & 2 & 5.25 \\ \hline
		\multirow{3}{*}{CenDE-DOBL} & mean & 98.67 & 98.68 & 78.79 & 81.90 & 90.95 & 88.39 &  \\
		& stddev & 2.81 & 1.08 & 8.64 & 3.17 & 10.15 & 5.09 &  \\
		& rank & 1 & 1 & 1 & 1 & 1.5 & 1 & 1.08 \\ \hline
	\end{tabular}
\end{table*}

In the last experiment, we compare our algorithm with 11 population-based trainers, namely particle swarm optimisation~\cite{ANN_PSO01}, artificial bee colony (ABC)~\cite{ANN_ABC01}, imperialist competitive algorithm (ICA)~\cite{ANN_ICA01}, firefly algorithm (FA)~\cite{ANN_FA01}, grey wolf optimiser (GWO)~\cite{ANN_GWO01}, ant lion optimiser~\cite{ANN_ALO01}, dragonfly algorithm (DA)~\cite{ANN_DA01}, sine cosine algorithm~\cite{ANN_SCA01}, whale optimisation algorithm (WOA)~\cite{ANN_WOA01}, grasshopper optimisation algorithm~\cite{ANN_GOA01}, and salp swarm algorithm (SSA)~\cite{SSA_01}. Algorithms such as PSO and ABC are among established training algorithms, while some others such as GOA and WOA are more recent. 

The results are reported in Table~\ref{Result_pop} from where it is evident that our proposed algorithm gives the best results on all datasets, providing clearly better performance compared to all other PBMHs.

\begin{table*}[]
	\centering
	\caption{10CV classification accuracy for all datasets for population-based training algorithms in comparison to CenDE-DOBL.}
	\label{Result_pop}
	\begin{tabular}{l|c|cccccc|c}
		&  & Iris & Cancer & Liver & Pima & Seed & Vertebral & avg. rank \\ \hline
		\multirow{3}{*}{PSO} & mean & 96.00 & 97.95 & 73.36 & 77.60 & 78.10 & 86.45 &  \\
		& stddev & 5.62 & 1.72 & 6.28 & 3.24 & 11.92 & 8.02 &  \\
		& rank & 5 & 6 & 3 & 8 & 6 & 3 & 5.17 \\ \hline
		\multirow{3}{*}{ABC} & mean & 84.67 & 97.95 & 70.75 & 78.26 & 72.38 & 82.90 &  \\
		& stddev & 9.45 & 1.03 & 6.47 & 4.45 & 8.03 & 5.70 &  \\
		& rank & 12 & 5 & 7 & 5 & 8.5 & 7 & 7.42 \\ \hline
		\multirow{3}{*}{ICA} & mean & 96.67 & 97.22 & 72.39 & 79.42 & 84.76 & 86.77 &  \\
		& stddev & 4.71 & 1.46 & 11.99 & 5.77 & 10.24 & 4.67 &  \\
		& rank & 4.00 & 10.00 & 5.00 & 2.00 & 2.00 & 2.00 & 4.17 \\ \hline
		\multirow{3}{*}{FA} & mean & 92.00 & 97.66 & 73.55 & 78.90 & 72.38 & 85.81 &  \\
		& stddev & 5.26 & 1.97 & 12.64 & 4.35 & 14.69 & 6.12 &  \\
		& rank & 9 & 8 & 2 & 4 & 8.5 & 4.5 & 6.00 \\ \hline
		\multirow{3}{*}{GWO} & mean & 93.33 & 98.10 & 73.01 & 67.45 & 78.10 & 81.94 &  \\
		& stddev & 4.44 & 1.39 & 9.74 & 2.79 & 10.09 & 7.93 &  \\
		& rank & 7 & 2 & 4 & 12 & 5 & 10.5 & 6.75 \\ \hline
		\multirow{3}{*}{ALO} & mean & 94.67 & 98.10 & 71.06 & 78.12 & 80.48 & 85.48 &  \\
		& stddev & 2.81 & 0.99 & 6.20 & 5.89 & 8.53 & 4.37 &  \\
		& rank & 6 & 3 & 6 & 6 & 4 & 6 & 5.17 \\ \hline
		\multirow{3}{*}{DA} & mean & 92.67 & 97.51 & 70.42 & 77.85 & 70.48 & 81.94 &  \\
		& stddev & 5.84 & 1.83 & 7.01 & 5.40 & 7.03 & 5.31 &  \\
		& rank & 8 & 9 & 9 & 7 & 11.5 & 10.5 & 9.17 \\ \hline
		\multirow{3}{*}{SCA} & mean & 90.67 & 97.08 & 65.50 & 74.47 & 71.43 & 82.26 &  \\
		& stddev & 7.83 & 1.82 & 5.96 & 4.20 & 8.98 & 10.67 &  \\
		& rank & 10 & 11 & 11 & 11 & 10 & 9 & 10.33 \\ \hline
		\multirow{3}{*}{WOA} & mean & 87.33 & 97.07 & 62.87 & 76.95 & 70.48 & 79.03 &  \\
		& stddev & 8.58 & 1.96 & 6.40 & 3.65 & 8.92 & 10.99 &  \\
		& rank & 11 & 12 & 12 & 10 & 11.5 & 12 & 11.42 \\ \hline
		\multirow{3}{*}{GOA} & mean & 98.00 & 98.09 & 70.73 & 79.03 & 84.29 & 82.58 &  \\
		& stddev & 3.22 & 1.84 & 6.45 & 3.72 & 12.10 & 6.49 &  \\
		& rank & 2.5 & 4 & 8 & 3 & 3 & 8 & 4.75 \\ \hline
		\multirow{3}{*}{SSA} & mean & 98.00 & 97.80 & 69.85 & 77.34 & 77.62 & 85.81 &  \\
		& stddev & 3.22 & 2.42 & 7.78 & 6.50 & 9.27 & 7.16 &  \\
		& rank & 2.5 & 7 & 10 & 9 & 7 & 4.5 & 6.67 \\ \hline
		\multirow{3}{*}{CenDE-DOBL} & mean & 98.67 & 98.68 & 78.79 & 81.90 & 90.95 & 88.39 &  \\
		& stddev & 2.81 & 1.08 & 8.64 & 3.17 & 10.15 & 5.09 &  \\
		& rank & 1 & 1 & 1 & 1 & 1 & 1 & 1.00 \\ \hline
	\end{tabular}
\end{table*} 

Overall, our proposed CenDE-DOBL algorithm thus gives excellent performance in comparison to the other 26 training algorithms.

\section{Conclusions} 
\label{sec:concl}
Training plays a crucial role in the performance of multi-layer neural networks. Conventional algorithms such as back-propagation are extensively employed in the literature, but suffer from difficulties such as their tendency to get stuck in local optima.

In this paper, we have proposed a novel differential evolution-based training algorithm, CenDE-DOBL, to find optimal weights in multi-layer neural networks. Our proposed algorithm benefits from a centroid-based strategy where a centroid individual is injected into the current population, and employs opposition-based learning in two ways, during initialisation and for generating jumps, while for further improvement a DE/local-to-best/1 strategy is used for mutation. Extensive experiments on diverse classification problems and in comparison to 26 conventional and population-based training algorithms, convincingly demonstrate CenDE-DOBL to yield excellent performance.  

In future, we intend to extend our approach to other types of neural networks including deep belief networks (DBNNs). Since DBNNs have a plethora of weights, the algorithm will need to be adapted to tackle this. In addition, our algorithm can be extended to optimise both weights and network structure simultaneously.

\bibliography{jalal}
\bibliographystyle{IEEEtran}

\end{document}